\documentclass{article}

\usepackage{arxiv}

\usepackage[utf8]{inputenc} % allow utf-8 input
\usepackage[T1]{fontenc}    % use 8-bit T1 fonts
\usepackage{hyperref}       % hyperlinks
\usepackage{url}            % simple URL typesetting
\usepackage{booktabs}       % professional-quality tables
\usepackage{amsfonts}       % blackboard math symbols
\usepackage{nicefrac}       % compact symbols for 1/2, etc.
\usepackage{microtype}      % microtypography
\usepackage{lipsum}		% Can be removed after putting your text content
\usepackage{graphicx}
\usepackage{natbib}
\usepackage{doi}
\usepackage{subcaption}
\usepackage{longtable} % para tabelas longas
\usepackage{array} % para formatação de colunas
\usepackage{geometry} % para ajustar as margens
\usepackage{url}
\usepackage{csquotes}
\usepackage{amsmath}
\usepackage{hyperref}

\usepackage[para,online,flushleft]{threeparttable}

\title{Análise de ambiguidade linguística em modelos de linguagem de grande escala (LLMs)}

\author{ \href{https://orcid.org/0000-0002-4934-8505}{\includegraphics[scale=0.06]{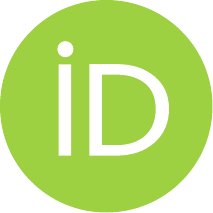}\hspace{1mm}Lavínia de Carvalho Moraes}\thanks{Agradecimento: processo nº 2022/14054-4, Fundação de Amparo à Pesquisa do Estado de São Paulo (FAPESP)--- Agradecimento: processo nº 2022/16103-2, Financiadora de Estudos e Projetos (FINEP).} \\
	Attenty Sistemas de Software \\
	Campinas, SP 13025-270 \\
	\texttt{l237294@dac.unicamp.br} \\
	\And
	\href{https://orcid.org/0000-0002-7737-7469}{\includegraphics[scale=0.06]{orcid.pdf}\hspace{1mm}Irene Cristina Silvério} \\
	Attenty Sistemas de Software \\
	Campinas, SP 13025-270 \\
	\texttt{i169329@dac.unicamp.br} \\
    \AND
      Rafael Alexandre Sousa Marques \\
	 Attenty Sistemas de Software \\
	 Campinas, SP 13025-270 \\
	 \texttt{rafaelalexandre001@gmail.com} \\
	 \And
	 Bianca de Castro Anaia \\
	 Attenty Sistemas de Software \\
	 Campinas, SP 13025-270 \\
	 \texttt{biancaanaia@gmail.com} \\
	 \And
	 Dandara Freitas de Paula \\
	 Attenty Sistemas de Software \\
	 Campinas, SP 13025-270 \\
	 \texttt{d206236@dac.unicamp.br} \\
      \And
	 Maria Carolina Schincariol de Faria \\
	 Attenty Sistemas de Software \\
	 Campinas, SP 13025-270 \\
	 \texttt{mariaschincariol@outlook.com} \\
      \And
	 Iury Cleveston \\
	 Attenty Sistemas de Software \\
	 Campinas, SP 13025-270 \\
	 \texttt{iury@attenty.com.br} \\
      \And
	 Alana de Santana Correia \\
	 Attenty Sistemas de Software \\
	 Campinas, SP 13025-270 \\
	 \texttt{alana@attenty.com.br} \\
    \And
	 Raquel Meister Ko Freitag \\
      Departamento de Letras Vernáculas (DLEV) \\
	 Universidade Federal de Sergipe \\
	 Sergipe, SE 49107-230 \\
	 \texttt{rkofreitag@uol.com.br} \\
}

\renewcommand{\shorttitle}{\textit{arXiv} Template}

\hypersetup{
pdftitle={A template for the arxiv style},
pdfsubject={q-bio.NC, q-bio.QM},
pdfauthor={David S.~Hippocampus, Elias D.~Striatum},
pdfkeywords={First keyword, Second keyword, More},
}

\begin{document}
\maketitle

\begin{abstract}
A ambiguidade linguística ainda é um grande desafio para sistemas de processamento de linguagem natural (NLP) apesar dos avanços em arquiteturas como Transformers e BERT. Inspirado pelo êxito recente dos modelos instrucionais ChatGPT (versão 3.5) e Gemini (Em 2023, a inteligência artificial chamava-se Bard), este trabalho visa analisar e discutir a ambiguidade linguística nesses modelos a partir de três tipos de ambiguidade no Português Brasileiro: semântica, sintática e lexical. Para isso, foi desenvolvido um corpus com 120 frases ambíguas e não ambíguas, submetidas aos modelos para tipificação, explicação e desambiguação. Também foi explorada a capacidade de geração de frases ambíguas, solicitando a geração de conjuntos de frases para cada tipo de ambiguidade. Os resultados foram analisados qualitativamente, com base em referenciais linguísticos reconhecidos, e quantitativamente pela acurácia das respostas obtidas. Evidenciamos que, mesmo os modelos mais sofisticados, como ChatGPT e Gemini, persistem equívocos e deficiências em suas respostas, com explicações frequentemente inconsistentes. A acurácia foi de no máximo 49,58\%, apontando a necessidade de estudos descritivos para o aprendizado supervisionado.
\end{abstract}

% keywords can be removed
\keywords{ambiguidade \and modelos de linguagem \and ChatGPT \and Gemini}

\section{Introdução}\label{sec:intro}

A ambiguidade linguística, caracterizada pela possibilidade de uma palavra ou frase ter dois ou mais significados distintos em uma sentença, é um fenômeno complexo para os modelos de linguagem natural \cite{ortega2023linguistic}. Essa complexidade deriva da riqueza e sutilezas inerentes à estrutura e ao uso das línguas humanas, a partir da multiplicidade de significados que palavras e estruturas linguísticas podem assumir, dependendo do contexto em que são utilizadas. Mesmo os modelos de linguagem mais avançados, tais como Transformer\cite{vaswani2017attention} e BERT\cite{devlin2018bert}, enfrentam desafios ao lidar com diversos tipos de ambiguidade, devido à necessidade de considerar uma ampla variedade de contextos, conhecimentos prévios e nuances culturais que influenciam a interpretação das palavras e frases. Discernir o significado correto em um contexto específico demanda não apenas uma compreensão profunda da língua, mas também uma capacidade de inferência e abstração historicamente desafiadoras de replicar em sistemas computacionais \cite{ortega2023linguistic}. %Além disso, os modelos frequentemente apresentam dificuldades ao identificar e desambiguar um texto em situações mais complexas, nas quais múltiplas camadas de interpretação estão presentes.

A partir de 2010, a área de NLP testemunhou um grande avanço tecnológico com a evolução de técnicas de deep learning, principalmente em funções como sumarização de texto, classificação de tópicos, análise de sentimentos e sintetização de voz. Assim, foi possível experienciar tarefas com uma qualidade que ainda não era esperada até o momento.
Desde o lançamento do ChatGPT e do Gemini\footnote[1]{O termo ChatPGT e Gemini serão tratados ao longo deste trabalho com pronomes masculinos por ser uma convenção adotada pelas demais literaturas brasileiras.}, modelos de linguagem instrucionais que têm revolucionado o mercado desde 2022, em consonância com o avanço tecnológico e o aumento da produtividade em diversas áreas do cotidiano, a inteligência artificial generativa tem impactado significativamente diversos setores, desde a educação até o mercado de trabalho. Essa influência se manifesta na distribuição de informações, na comunicação de ideias e na compreensão de dados discursivos, redefinindo a forma como interagimos e lidamos com o vasto espectro de informações disponíveis. Apesar dos avanços, subsistem incertezas se a ChatGPT e a Gemini conseguem compreender fenômenos linguísticos complexos com a mesma facilidade que os seres humanos. Ainda há limitações quanto à capacidade desses modelos em processarem adequadamente as ambiguidades linguísticas, bem como se conseguem superar os desafios históricos observados em modelos tradicionais, assim como suas limitações e pontos fortes. Os poucos estudos desenvolvidos têm como foco na língua inglesa \cite{ortega2023linguistic}, análises no contexto do português brasileiro, uma língua de baixos recursos linguísticos \cite{finger2021inteligencia}, ainda não foram conduzidas, revelando a importância de investigações neste campo.

Nesse contexto, este trabalho tem como objetivo responder as seguintes perguntas:

\begin{enumerate}
    \item Qual é a precisão dos modelos na detecção de ambiguidade linguística em frases do Português Brasileiro?    
    \item Os modelos conseguem desambiguar adequadamente as sentenças?    
    \item Qual dos modelos percebe melhor os fenômenos de homonímia e polissemia?    
    \item Quais padrões de ambiguidade os modelos ChatGPT e Gemini demonstram conhecer na
    geração de frases ambíguas?
\end{enumerate}

O estudo de ambiguidade é particularmente complexo, por envolver uma gama de variáveis que interferem, desde a natureza do item lexical, passando pela sintaxe da sentença, e envolvendo o conhecimento de mundo e experiência pessoal de cada falante. Além disso,  há um paradoxo a ser superado: um humano pode perceber ambiguidade e não saber explicá-la, enquanto a IA pode saber explicar o que é uma ambiguidade, mas não saber reconhecê-la.  
Assumindo que as IAs são capazes de imitar em grande medida o processamento da linguagem humana e que têm o potencial de fornecer informações sobre a forma como as pessoas aprendem e utilizam a linguagem \cite{cai2023does} conduzimos um estudo com a realização de quatro tarefas. Utilizando um conjunto de sentenças com e sem ambiguidade criado por nós. Verificamos a consistência das respostas dos modelos ao fazer as mesmas perguntas duas vezes, em momentos distintos, e contrastamos algumas inconsistências nas respostas obtidas. 
\section{Processamento da ambiguidade linguística e o processamento da linguagem natural}\label{sec-referencial-teorico}

O processamento semântico é um domínio de investigação fundamental para os modelos linguísticos pré-treinados e dos grandes modelos de linguagem. No campo das tarefas de processamento semântico, a desambiguação do sentido das palavras demanda parâmetros definidos. No entanto, o estudo da ambiguidade é multidimensional na linguística, como veremos a seguir.

Esta pesquisa fundamenta-se em disciplinas que estão interligadas, demonstrando a natureza interdisciplinar das áreas envolvidas. Portanto, o estudo aproveita-se dos princípios teóricos relacionados à ambiguidade linguística (Seção \ref{sec-ambiguidade-linguistica}) e da aplicação dos modelos computacionais de grande escala, ChatGPT e Gemini (Seção \ref{sec-modelos-linguagem}).

É importante destacar que o arcabouço teórico sobre ambiguidade linguística, empregado nesta seção, servirá como alicerce para a construção do nosso conjunto de sentenças. Entre várias abordagens teóricas disponíveis, decidimos adotar a perspectiva de Cançado \cite{canccado2005manual}, dada sua abrangência em capturar em apenas uma taxonomia uma ampla gama de aspectos fundamentais da ambiguidade linguística, além do seu amplo reconhecimento no âmbito acadêmico.

\subsection{Ambiguidade Linguística}\label{sec-ambiguidade-linguistica}

A ambiguidade é um fenômeno linguístico-cognitivo universal no qual uma palavra ou uma sentença podem apresentar o desencadeamento de mais de uma interpretação plausível. Neste aspecto, a ambiguidade só pode ser resolvida através do contexto, de forma que a contextualização tem a função de estabelecer qual dos possíveis sentidos será selecionado. 
Os estudos sobre a ambiguidade costumam considerar o nível de uso da língua. Seguimos a proposta de Cançado (2005) para o português brasileiro, com a ambiguidade lexical, dividida em casos de homonimia e polissemia, ambiguidade semântica e a ambiguidade sintática.

A \textbf{ambiguidade lexical} compreende uma sentença com  dupla interpretação incidente em um item lexical, podendo ser gerada por homonímia ou polissemia. A homonímia se dá quando os sentidos do item lexical não são relacionados, como na oração ``Eu estou indo para o banco'' em que a palavra ``banco'' possui significados diferentes, pode corresponder à instituição financeira e ao assento. Já a polissemia ocorre quando os sentidos do termo identificado como ambíguo contém relação entre si, por exemplo na sentença ``O Frederico esqueceu a sua concha'', neste caso ``concha'' pode significar uma concha do mar ou uma concha de cozinha, em que ambos os objetos possuem o mesmo formato, e por isso, pode ocorrer uma associação polissêmica, o mesmo fenômeno ocorre com palavras como: rede (de internet, de deitar, de pescar) e pilha (de comida, de bateria).

Nesta taxonomia, a \textbf{ambiguidade semântica} é abordada como uma questão de correferencialidade, em que os pronomes podem ter vários antecedentes. Consideremos, por exemplo, a seguinte frase: ``José falou com seu irmão?'' Esta sentença ilustra claramente esse tipo de ambiguidade, na qual não é possível determinar se o irmão mencionado é o irmão de José ou o irmão da pessoa para quem a pergunta é dirigida, ou seja, um terceiro elemento. Nesse contexto, as interpretações possíveis são atribuídas à natureza da ligação entre os pronomes presentes na sentença.

A \textbf{ambiguidade sintática} é um fenômeno de imprecisão de sentidos que não é ocasionado pela interpretação de uma palavra individual, mas se atribui às distintas estruturas sintáticas que originam diferentes interpretações: a frase concebe diferentes análises a partir dos seus possíveis sintagmas, que são divisões existentes dentro da frase em grupos de palavras. A sentença ``O magistrado julga as pessoas culpadas'' é um exemplo em que a organização da sentença pode ser: (a) O magistrado [julga] [as crianças culpadas] ou (b) O magistrado [julga] [culpadas] [as crianças]. A ambiguidade sintática envolve as diversas possibilidades de interpretação da sentença apenas reorganizando a posição das expressões envolvidas na oração, o que não ocorre nos tipos de ambiguidade tratados anteriormente. A ambiguidade sintática é um fenômeno amplamente estudado, com padrões obedecendo a princípios como o princípio da aposição mínima e da aposição local \cite{maiadimensoes}, que atuam em diferentes línguas, incluindo o português brasileiro \cite{maia2003processamento,maia2004compreensao,brito2013processamento, machado1996sintaxe}. A ambiguidade semântica, ou anafórica, também apresenta padrões regulares na retomada de antecedentes a pronomes \cite{bruscato2021resoluccao,nogueira2014resoluccao, godoy2020efeitos, de2023interpretaccao}. Para o português, a ambiguidade lexical é a que apresenta mais recursos descritivos para suporte computacional \cite{laporte2001resoluccao}, necessário aos LLMs.

\subsection{Modelos de Linguagem}\label{sec-modelos-linguagem}

Os modelos de linguagem grandes (LLMs), como o ChatGPT 3.5 \cite{openai2023gpt4} e o Gemini \cite{ahmed2023chatgpt}, contam com aproximadamente 175 bilhões e 1,5 trilhão de parâmetros. Eles funcionam a partir de redes atencionais do tipo Transformer \cite{vaswani2017attention} que são pré-treinadas de forma auto-superivisionada em grandes conjuntos de dados. Posteriormente, eles são refinados através do aprendizado instrucional baseado em contexto e através do aprendizado por reforço baseado em feedback humano (RLHF) \cite{ouyang2022training}. Os mecanismos atencionais do tipo \textit{self-attention} \cite{vaswani2017attention} presentes nas arquiteturas permitem a captura de dependências de longas distâncias de forma computacionalmente eficaz, minimizando o esquecimento dos modelos em sequências longas. Por fim, a estratégia de indução de pensamento em cadeia (CoT) \cite{wei2023chainofthought} aplicada após os treinamentos permite que os modelos usem a sua última saída como entrada para gerar uma saída ainda mais refinada, melhorando a qualidade das respostas dadas.

No contexto da ambiguidade linguística, sabe-se que os mecanismos de \textit{self-attention} aprimoram a capacidade dos modelos em lidar com a ambiguidade semântica por meio do aprendizado da correferenciação, naturalmente presente na estrutura do mecanismo atencional \cite{ortega2023linguistic}. Entretanto, ainda não há evidências de que o RLHF, CoT e treinamentos instrucionais exclusivamente  presentes na ChatGPT e Gemini impactam no processamento da ambiguidade.

Assumimos a hipótese de que o aprendizado instrucional permite, de forma implícita, que esses modelos compreendam instruções pelo contexto e sigam direções específicas sobre a intenção do usuário, mitigando ou resolvendo certos casos de ambiguidade. Além disso, o RLHF pode ser particularmente valioso para lidar com situações ambíguas, já que ele provê o alinhamento correto do modelo com as intenções do usuário, por meio de recompensas dadas durante um treinamento de ajuste fino. Por fim, é possível que o CoT também tenha um papel crucial na resolução de ambiguidade, pois pode auxiliar os modelos a decompor o problema em etapas intermediárias mais gerenciáveis, permitindo abordar a resolução de ambiguidade de forma gradual, em vez de tentar resolvê-la de uma só vez. Diante dessas considerações, surge a questão: como esses elementos se adaptam durante o treinamento dos modelos para lidar com as sutilezas e complexidades do processamento de ambiguidades no contexto do português brasileiro? Explorar essa questão pode fornecer informações valiosas sobre como otimizar esses modelos para atender às nosssas necessidades linguísticas e culturais.
\section{Metodologia}\label{sec-metodologia}

Para avaliar o desempenho dos modelos no processamento de ambiguidades quanto aos parâmetros da ambiguidade lexical, sintática e semântica, conduzimos tarefas utilizando um conjunto de dados composto por 120 sentenças seguindo a distribuição das sentenças ambíguas por um balanceamento quanto aos três tipos de ambiguidade. Dessas, 60 apresentam algum tipo de ambiguidade, seja semântica, lexical ou sintática. As outras 60 sentenças distratoras tiveram sua ambiguidade barrada ao máximo e não foi verificada por juízes humanos. Foram criadas 20 sentenças com ambiguidade lexical - compreende casos de homonímia e polissemia, mas sem distinção de categoria na análise das frases -, 20 sentenças com ambiguidade semântica na qual o referente dos pronomes não está claro. Por fim, 20 sentenças com ambiguidade sintática envolvendo adjuntos adnominais ou adverbiais ambíguos que provocam duplo sentido devido a diferentes organizações estruturais que a frase pode ter (Tabela \ref{tab:amostras_sentencas}). Todas as frases e análises dos resultados descritos foram realizadas por seis especialistas humanos do curso de Letras ou Linguística com conhecimento na área.

\begin{table}[htpb]
\centering
\begin{threeparttable}
\caption{}
\label{tab:amostras_sentencas}
\begin{tabular}{llp{6cm}}
\toprule
Id & Sentença & Classe \\
\midrule
1 & João foi à mangueira, assim como Maria. & Ambiguidade lexical \\
2 & Ela não gosta da amiga dela. & Ambiguidade semântica \\
3 & O menino viu o incêndio do prédio. & Ambiguidade sintática \\
\bottomrule
\end{tabular}
\url{https://osf.io/u7wre/?view_only=572c74eb4c634d47a02ad25485ea8caa}.
%\notes{Se necessário, poderá ser adicionada uma nota ao final da tabela.}
\end{threeparttable}
\end{table}

Para responder as nossas perguntas de pesquisa, foram conduzidas quatro tarefas distintas com as sentenças criadas. Em todas, foram realizadas coletas duplicadas das interações para cada frase, reiniciando o \textit{console} entre cada coleta para evitar qualquer influência do contexto que pudesse gerar respostas tendenciosas. Essa abordagem permitiu avaliar a consistência dos modelos nas respostas fornecidas.

A tarefa 1 visava indentificar se os modelos conseguem detectar a presença de ambiguidade em cada sentença por meio da seguinte instrução: \textbf{A sentença ``[sentença]'' é ambígua? Responda, sim, não ou não sei}. Foram apresentadas individualmente todas as sentenças e registradas as respostas dos modelos, comparando-as com a nossa classificação prévia. As respostas foram cuidadosamente avaliadas quanto à correção e abrangência das explicações fornecidas por 6 juízes-humanos que as julgaram independemente. A partir dos resultados, foi gerada uma matriz de confusão para computar a quantidade de verdadeiros positivos (sentenças que são ambíguas e que os modelos classificaram como ambíguas), falsos positivos (sentenças que não o são e que os modelos classificaram como ambíguas), verdadeiros negativos (sentenças que não são ambíguas e que os modelos assim classificaram como não ambíguas), e falsos negativos (sentenças que são ambíguas e que os modelos classificaram como não ambíguas).

Na tarefa 2, foi realizado um teste para avaliar a capacidade dos modelos em distinguir corretamente entre as três classes de ambiguidade estudadas neste trabalho, formulando a seguinte pergunta para cada modelo: \textbf{``Qual o tipo de ambiguidade?''}. A tarefa consistiu em perguntar qual o tipo de ambiguidade da sentença que foi classificada anteriormente como ambígua ou não ambígua. Na tarefa 3, foi verificada a capacidade dos modelos em desambiguar as sentenças que foram fornecidas a eles. Com esse propósito, foram apresentadas frases que incluem tanto sentenças ambíguas, quanto sentenças não ambíguas, e solicitado aos modelos a seguinte instrução: \textbf{Faça a desambiguação da frase: ``[sentença]''}. A tarefa busca testar a habilidade dos modelos em compreender e interpretar o contexto, escolhendo a interpretação mais apropriada quando a ambiguidade está presente. 

Na tarefa 4, foi avaliada a capacidade dos modelos em gerar frases ambíguas na categoria solicitada. Para isso, pedimos para cada modelo gerar frases da seguinte forma: \textbf{Gere 20 frases com ambiguidade ``[categoria]''}. Em seguida, as respostas obtidas foram avaliadas por juízes-humanos, buscando compreender quão precisas são a ChatGPT e o Gemini ao criarem frases que apresentam múltiplas interpretações contextuais.

Para mensurar quantitativamente o desempenho dos modelos, foi utilizada a métrica de acurácia, a qual já é amplamente empregada na área de aprendizado de máquina  \cite{naser2021error, freitag2021funccao}. A acurácia, no contexto da classificação, representa a proporção de frases corretamente classificadas pelos modelos em relação ao total de frases apresentadas na tarefa, como apresentado na equação \ref{eq_1}.

\begin{equation}
    \text{acc} = \frac{\text{Número de previsões corretas}}{\text{Total de previsões}}
    \label{eq_1}
\end{equation}

Todas as sentenças criadas por nós e geradas pelos modelos durante as tarefas estão disponíveis no Apêndice \ref{sec-apendice}. Todas as respostas dos modelos durante os tarefas assim como o nosso conjunto de sentenças está disponível para download\footnote{\url{https://osf.io/u7wre/?view_only=572c74eb4c634d47a02ad25485ea8caa}} em nosso repositório.
\section{Resultados}\label{sec-experimentos-resultados}

Nesta seção, são apresentados os resultados obtidos. Cada seção responde uma das nossas perguntas de pesquisa através de quatro tarefas. Todas as análises dos resultados descritos foram realizadas por seis especialistas humanos do curso de Letras ou Linguística com conhecimento na área.

\subsection{Qual é a precisão dos modelos na detecção de ambiguidade linguística em frases do Português Brasileiro?}\label{resultados-q-1}

Para avaliar a precisão dos modelos na detecção de ambiguidade seguimos a proposta de \cite{freitag2021funccao}, foram comparadas as acurácias e as matrizes de confusão obtidas, usando os dados da tarefa 1. Esta análise concentrou-se exclusivamente na detecção da presença ou ausência de ambiguidade, sem levar em consideração o tipo específico identificado pelos modelos posteriormente. Assim, os dados foram divididos em dois grupos distintos, totalizando 60 frases com ambiguidade e 60 frases distratoras. Foram realizadas duas coletas por frase, sendo obtidas 240 predições para cada modelo.

\begin{figure}[htb]
    \centering
    \begin{subfigure}[b]{0.45\textwidth}
        \includegraphics[width=\textwidth]{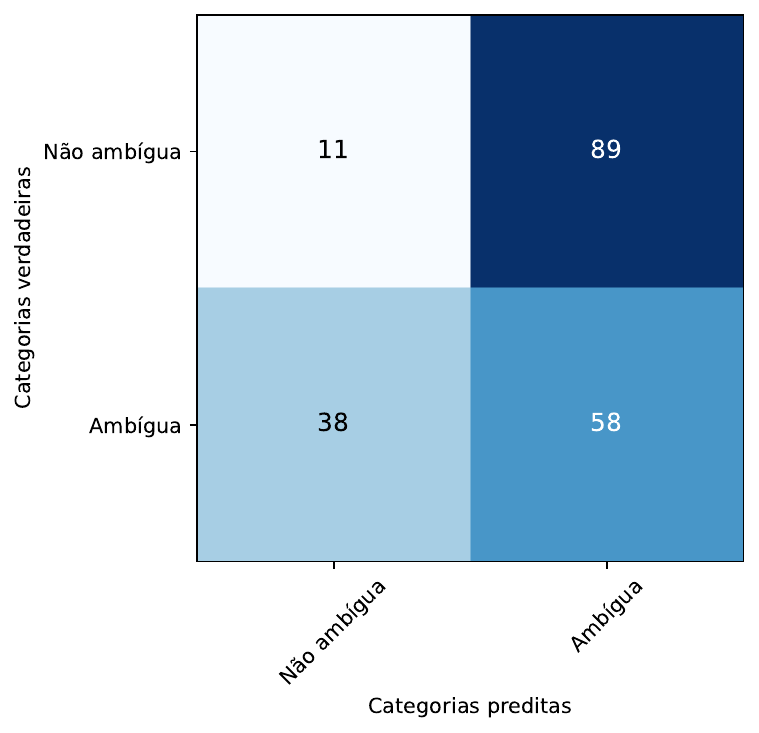}
        \caption{Matriz de confusão do ChatGPT.}
        \label{fig:matriz_confusao_chatgpt}
    \end{subfigure}
    \hfill
    \begin{subfigure}[b]{0.45\textwidth}
        \includegraphics[width=\textwidth]{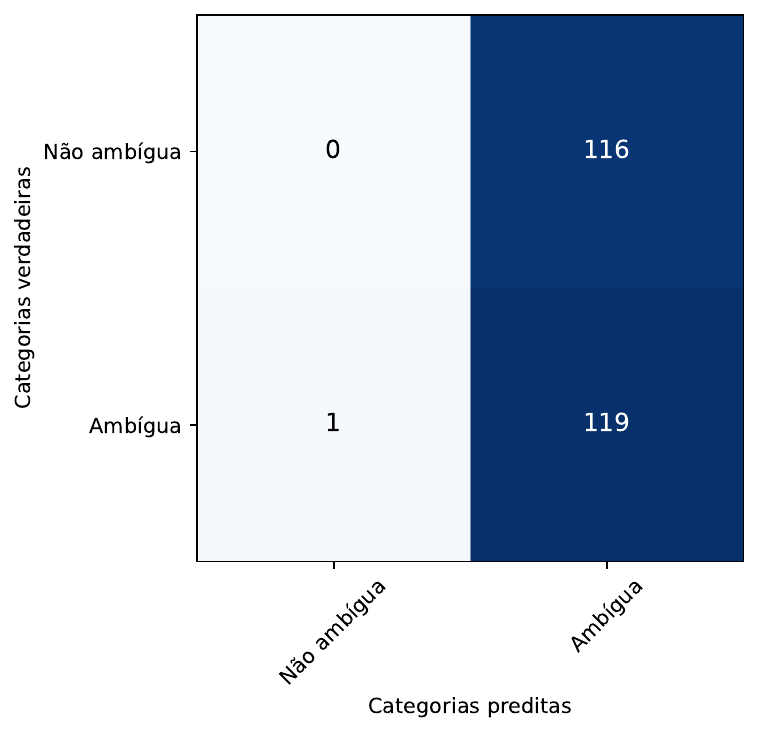}
        \caption{Matriz de confusão do Gemini.}
        \label{fig:matriz_confusao_bard}
    \end{subfigure}
    \caption{Matrizes de Confusão dos modelos ChatGPT e Gemini.}
    \label{fig:matriz_confusao_1}
\end{figure}

Os resultados da matriz de confusão na Figura \ref{fig:matriz_confusao_1} revelam que o ChatGPT registrou uma acurácia de apenas 28,75\%, enquanto o Gemini alcançou 49,58\%, indicando que estas versões dos modelos não conseguem detectar ambiguidade com precisão confiável. Os resultados revelam que ambos os modelos exibem uma quantidade significativa de falsos positivos, identificando ambiguidade em frases que carecem dela. Enquanto o ChatGPT demonstra erros distribuídos em todos os quadrantes da matriz, o Gemini tende a rotular quase todas as frases como ambíguas, resultando em uma taxa maior de falsos positivos. Para computar as matrizes de confusão e a acurácia foram consideradas apenas as respostas em que os modelos responderam \enquote{Sim} e \enquote{Não}, de modo que todas as respostas \enquote{Não Sei} foram descartadas. Assim, o ChatGPT destacou-se por apresentar mais dúvidas, conseguindo responder 196 perguntas enquanto o Gemini respondeu 236.

A diferença de acurácia entre os dois modelos pode ser atribuída ao fato do ChatGPT expressar dúvidas ao detectar frases ambíguas, declarando não saber ou negando a presença de ambiguidade. Analisando os três tipos de ambiguidade, observa-se que o ChatGPT lida melhor com ambiguidades semânticas e sintáticas, cometendo mais erros quando a ambiguidade é apenas lexical. Uma explicação é devida à estrutura da ambiguidade sintática ser descrita em estudos de processamento linguístico \cite{maiadimensoes} (1) e de processamento de linguagem natural \cite{padovani2022metodo}(2). Em contrapartida, o Gemini apresentou apenas um caso de falso negativo, acertando todos os outros testes em frases ambíguas. Entretanto, o Gemini tem a tendência de não distinguir entre frases ambíguas e não ambíguas, pois, em todos os testes, indica a presença de ambiguidade.

%\textbf{Sintática}.Dentre os 40 experimentos com ambiguidade feitos com o ChatGPT e Bard, a primeira inteligência constatou ambiguidade em 22 frases, mas por outro lado, o Gemini identificou integralmente as sentenças ambíguas ainda que nem todas estivessem com a classificação correta. No entanto, no experimento feito com as frases sem ambiguidade, o ChatGPT reconheceu alguma ambiguidade em 8 casos, alcançando uma acurácia final de 37,5\%, enquanto que o Gemini identificou ambiguidade em todos os testes, resultando, nessa etapa, 100\% de falsos positivos e chegando a uma acurácia de 50\%.  

\subsection{Qual dos modelos percebe melhor os fenômenos de homonímia e polissemia?}\label{resultados-q-4}

Neste estudo, foram examinadas as explicações fornecidas pelos modelos na tarefa 2, analisando apenas as respostas do conjunto de frases com ambiguidade lexical. Durante a análise, foi considerado que os modelos perceberam a homonímia e a polissemia através da explicação dada pelos modelos para justificar o tipo de ambiguidade identificada. Se a explicação dada pelos modelos foi referente a ambiguidade gerada devido aos diferentes significados que o item lexical pode assumir na frase, e se realmente os juízes-humanos enxergariam os diferentes significados do item lexical da mesma maneira, foi considerado acerto, caso contrário foi tido como um erro por parte dos modelos. 

Para demonstrar temos as frases \textbf{Isso não é legal!} e \textbf{A carteira foi danificada.} que foram classificadas como ambiguidade lexical de homonímia pelos juízes humanos. Tais frases foram testadas no ChatGPT de modo que a primeira frase foi considerada correta, pois recebeu, em um dos testes, a seguinte resposta \textbf{A frase \enquote{Isso não é legal!} pode apresentar ambiguidade de sentido, pois a palavra \enquote{legal} possui múltiplas interpretações, dependendo do contexto em que é usada.} e a segunda frase foi classificada como incorreta, em um dos testes, por apresentar uma explicação incoerente \textbf{A frase \enquote{a carteira foi danificada} pode ser considerada ambígua devido à ambiguidade estrutural. Isso ocorre porque não está claro se a carteira sofreu dano físico ou se está se referindo a uma carteira de identidade ou pertencente a alguém. Portanto, a ambiguidade está relacionada à interpretação da frase em termos de sua estrutura sintática.}

Ambos os modelos demonstraram uma boa compreensão dos fenômenos de homonímia e polissemia. A ChatGPT obteve sucesso em 75\% das vinte frases testadas, enquanto a BARD alcançou uma taxa de acerto de 80\%. Vale ressaltar a notável similaridade na forma como os modelos interpretaram esses fenômenos, uma vez que ambos obtiveram sucesso quase no mesmo subconjunto de frases.

\subsection{Os modelos conseguem desambiguar adequadamente as sentenças?}\label{resultados-q-3}

Nesta seção, a análise foi baseada nas respostas obtidas na tarefa 3, no qual foi solicitada a desambiguação de todas as sentenças. As respostas obtidas foram divididas em três categorias: 1) Correta ocorreu quando o modelo explica corretamente duas ou mais interpretações possíveis para a sentença, de forma semelhante a interpretação que um ser humano teria. No caso das sentenças sem ambiguidade, o modelo esclarece a ausência de ambiguidade e observa que não é necessário executar a tarefa solicitada; 2) Incompleta ocorreu quando o modelo apresentou pelo menos uma interpretação correta para a sentença, errando nas demais interpretações possíveis; e 3) Incorreta ocorreu quando o modelo apresentou opções de desambiguação sem nenhum sentido. 

\begin{figure}[htb]
    \centering
    \begin{subfigure}[b]{0.45\textwidth}
        \includegraphics[width=\textwidth]{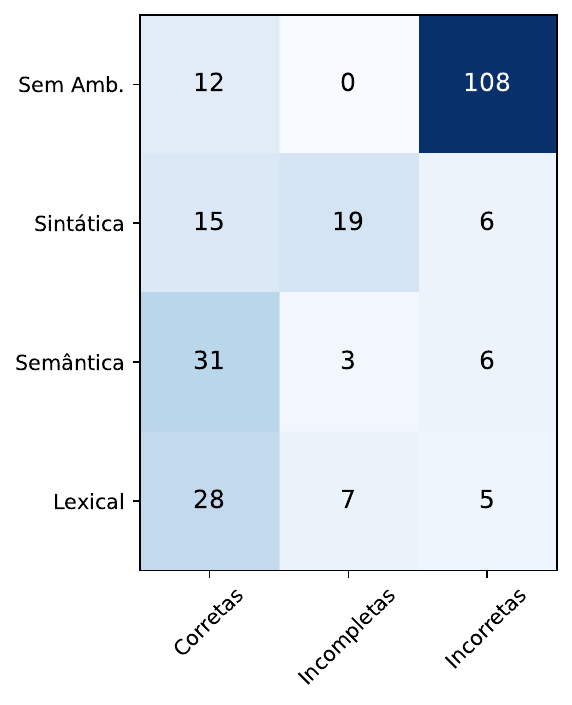}
        \caption{Resultados quantitativos da tarefa de desambiguação realizada pelo ChatGPT.}
        \label{fig:matriz_confusao_chatgpt}
    \end{subfigure}
    \hfill
    \begin{subfigure}[b]{0.45\textwidth}
        \includegraphics[width=\textwidth]{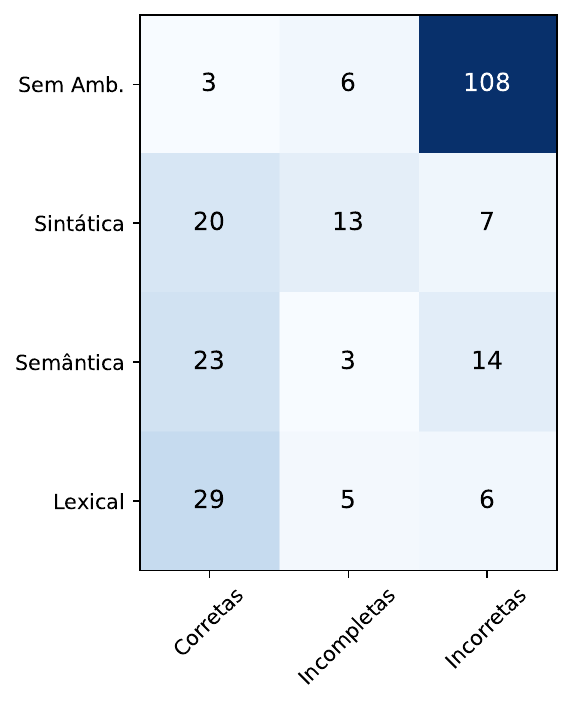}
        \caption{Resultados quantitativos da tarefa de desambiguação realizada pelo Gemini.}
        \label{fig:matriz_confusao_bard}
    \end{subfigure}
    \caption{Estatísticas obtidas na tarefa de desambiguação. Os resultados mostram que ambos os modelos apresentam muita dificuldade para identificar que as sentenças distratoras não tem ambiguidade. Em 108 casos de teste, ambos os modelos forneceram explicações que não refletiam a interpretação humana, justificando a presença de ambiguidade em sentenças que, na verdade, não a possuíam.}
    \label{fig:matriz-confusao-desamb}
\end{figure}

Os resultados mostrados na Figura \ref{fig:matriz-confusao-desamb} revelam que ambos os modelos enfrentam dificuldades ao tentar desambiguar corretamente as sentenças, uma vez que propõem interpretações equivocadas, identificando ambiguidade em elementos pouco ou nada relacionados à ambiguidade real. Por exemplo, no conjunto de frases sem ambiguidade, ambos os modelos tentaram desambiguá-las oferecendo opções de interpretação. 

A frase \textbf{Como o pássaro estava cansado, Ricardo o alimentou.} obteve uma desambiguação incorreta tanto na tarefa do ChatGPT como do Gemini. No primeiro modelo, a resposta foi a seguinte: \textbf{\enquote{Para desambiguar a frase  e torná-la mais clara, podemos reescrevê-la de forma a eliminar a ambiguidade de referência. Uma maneira de fazer isso é substituindo o pronome \enquote{o} por seu referente explícito. Vejamos: 1. Como Ricardo estava cansado, o pássaro foi alimentado por ele. Nesta versão reescrita, fica claro que o pronome \enquote{o} se refere ao pássaro, e não a Ricardo. Agora a frase não possui mais ambiguidade de referência.}} 

O segundo modelo apresentou a seguinte explicação: \textbf{\enquote{Claro, aqui estão duas possíveis desambiguações da frase \enquote{Como o pássaro estava cansado, Ricardo o alimentou}: Ricardo alimentou o pássaro porque estava cansado. Ricardo alimentou o pássaro porque estava cansado de ver o pássaro cansado. A primeira interpretação é a mais provável, pois é mais comum alimentar alguém quando está cansado. No entanto, a segunda interpretação também é possível, pois é possível que Ricardo estivesse preocupado com o bem-estar do pássaro e quisesse ajudá-lo a se recuperar. Sem mais informações, não é possível dizer com certeza qual é a interpretação correta.}}

Os modelos desambiguaram sem necessidade mais de 90\% dos casos de teste do conjunto de frases sem ambiguidade. É relevante destacar que, mesmo o ChatGPT - que afirmou em perguntas anteriores que algumas frases não tinham ambiguidade - ofereceu opções de desambiguação, sendo incongruente com suas respostas anteriores. Uma exceção notável ocorreu em uma resposta relacionada à frase \enquote{Gostamos da cor rosa}, na qual o ChatGPT afirmou: \textbf{\enquote{Peço desculpas, mas a frase 'Gostamos da cor rosa' não é intrinsecamente ambígua e, portanto, não requer desambiguação. Ela expressa claramente uma preferência pela cor rosa. Se houver um contexto específico causando ambiguidade, por favor, forneça mais informações para que eu possa ajudar a esclarecer a situação}}. Observa-se que os modelos têm a tendência de realizar superinterpretações das sentenças sem ambiguidade, oferecendo soluções sem lógica apenas para cumprir a tarefa solicitada, demonstrando ainda não haver o conhecimento explícito de regras do funcionamento da lígua.

%\begin{table}[h]
%\caption{Estatísticas obtidas na tarefa de desambiguação.}
%\label{tab:estat_desambiguacao}
%\begin{tabular}{lcccccc}
%\hline
%Grupo                 & \multicolumn{2}{c}{Corretas} & \multicolumn{2}{c}%{Incompletas} & \multicolumn{2}{c}{Incorretas} \\ \hline
%                      & ChatGPT        & BARD        & ChatGPT          & BARD         & ChatGPT         & BARD         \\
%Sem ambiguidade       & 12               & 3            & 0                 & 6             & 108                & 108             \\
%Ambiguidade Sintática & 15             & 20              & 19               & 13             & 6               & 7              \\
%Ambiguidade Semântica & 31               & 23              & 3                 & 3             & 6                & 14             \\
%Ambiguidade Lexical   & 28               & 29             & 7                 & 5           & 5                 & 6              \\ \hline
%\end{tabular}
%\source{Própria.}
%\notes{Se necessário, poderá ser adicionada uma nota ao final da tabela.}
%\end{table}

Na desambiguação das frases com ambiguidade sintática, os modelos também apresentaram erros na maioria dos testes. Em 62,5\% dos testes, as análises apresentadas estavam incorretas ou incompletas e os critérios usados nas frases classificadas com ambiguidade sintática, também, pendiam para uma abordagem semântica, lexical ou voltada para a necessidade de um contexto favorável para desambiguar as sentenças. No caso da sentença \enquote{Pedro me mandou um cartão postal de Veneza}, o ChatGPT apresentou a seguinte desambiguação para a frase: \enquote{Pedro me mandou um cartão postal de Veneza}; \enquote{Pedro mandou um cartão postal de Veneza para a minha amiga Maria, e ela me mostrou o cartão}. Contudo, a ambiguidade no adjunto \enquote{de Veneza} permanece. Outra solução proposta é mudar a alocação do pronome: \enquote{Pedro mandou um cartão postal de Veneza para mim}, mantendo a frase ambígua. Nesse caso, o Gemini ofereceu melhores opções, como \enquote{Pedro está atualmente em Veneza e me enviou um cartão postal que foi produzido lá.} ou \enquote{Pedro não está atualmente em Veneza, mas me enviou um cartão postal que mostra uma imagem da cidade.}, demonstrando aplicar os princípios da aposição mínima e da aposição local \cite{maiadimensoes} para resolução da ambiguidade. Embora o modelo da Google não consiga, na maioria das vezes, precisar qual é o elemento causador de ambiguidade.

Por outro lado, no grupo de frases com ambiguidade semântica, na maior parte dos testes, ambos os modelos de linguagem sugeriram frases adequadas para a desambiguação das sentenças, sempre destacando a importância do contexto para a correta interpretação. Um exemplo pode ser visto com a frase \enquote{Ana me contou um segredo sobre ela}, em que o ChatGPT propôs a desambiguação esperada, compreendendo que \enquote{ela} pode se referir tanto à Ana, quanto a uma outra pessoa. Em relação às respostas do Gemini, a maioria também foi apropriada porém com precisão menor que a ChatGPT. Por exemplo, em \enquote{Paulo não entrou na universidade de novo}, o Gemini indicou corretamente que as possíveis reescritas da sentença seriam \enquote{Paulo não foi aceito na universidade de novo} e \enquote{Paulo não entrou na universidade de novo depois de ter se formado}. Já em \enquote{A carteira foi danificada}, o Gemini sugeriu opções que especificassem informações não necessariamente ambíguas, visto que ele interpretou que a ambiguidade estaria em \enquote{danificado}, assim, deveria ser especificado se \enquote{A carteira foi danificada fisicamente} ou \enquote{A carteira foi danificada financeiramente}. Estas interpretações são bastante distantes do que se encontra em estudos psicolinguísticos sobre processamento \cite{machado1996sintaxe, brito2013processamento}.

Os modelos tiveram, simultaneamente, melhor desempenho nos casos de ambiguidade lexical (Figura \ref{fig:matriz-confusao-desamb}), entranto, o ChatGPT não conseguiu desambiguar corretamente algumas frases que envolvem polissemia. Na frase \enquote{João quer mangas, assim como Maria.}, o modelo apresentou três opções de desambiguação: \enquote{João quer mangas, da mesma forma que Maria quer.}; \enquote{João quer mangas da mesma forma que Maria quer mangas.} e, por fim, \enquote{João quer mangas e também quer Maria.}. Percebe-se, então, que o modelo tende a encontrar problemas na estrutura, e não nas palavras isoladamente. O Gemini apresentou mais facilidade de identificação e explicação nos casos de homonímias, haja vista que todas as suas explicações nestes casos estão corretas. Por outro lado, também predominaram justificativas equivocadas em alguns casos de polissemia.
\subsection{Quais padrões de ambiguidade os modelos ChatGPT e Bard demonstram conhecer na geração de frases ambíguas?}
\label{resultados-q-5}

A investigação sobre os padrões de ambiguidade usados pelos modelos ChatGPT e Gemini durante a geração de frases ambíguas é abordada por meio da análise das respostas obtidas na tarefa 4. Para conduzir essa análise, seis juízes-especialistas altamente qualificados julgaram se as frases geradas pelos modelos continham elementos que induzem ambiguidade perceptível por seres humanos e, em caso afirmativo, se esses elementos se alinhavam corretamente com a categoria de ambiguidade.

\textbf{Ambiguidade Lexical.} Na geração de frases com ambiguidade lexical, tanto o ChatGPT quanto o Gemini revelaram não conhecer os padrões geradores desse tipo de ambiguidade, resultando, em sua maioria, em frases sem qualquer forma de ambiguidade identificada pelos avaliadores humanos. O ChatGPT, em particular, não gerou nenhuma frase com ambiguidade lexical, tendo 18 frases avaliadas por especialistas sem qualquer notificação de ambiguidade. Por exemplo, foram geradas frases como \enquote{O pássaro voou em direção à árvore mais alta.} e \enquote{Ela encontrou uma bela maçã na floresta.}, em que para um observador humano não há ambiguidade, revelando clareza na expressão. Em apenas duas das frases geradas, foram identificados padrões de adjunto ambíguo, referindo-se a ambiguidade estrutural ou sintática, não se enquadrando corretamente na categoria de ambiguidade lexical.

O Gemini também enfrentou dificuldades ao gerar frases com ambiguidade lexical. Das 20 frases, em 13 delas nenhum avaliador humano conseguiu identificar qualquer tipo de ambiguidade, resultando em uma taxa de sucesso de apenas 15\% para o Gemini e de 0\% para o ChatGPT. Um exemplo é a frase \enquote{O advogado defendeu o criminoso.} que não apresenta ambiguidade perceptível. Foram observados apenas três casos em que a ambiguidade residia na homonímia e polissemia, como na frase \enquote{O professor ensinou a classe.}. Além disso, o Gemini, em alguns casos, misturou padrões que se alinhariam mais com a ambiguidade situacional do que com a ambiguidade lexical, por exemplo ao sugerir a frase \enquote{A casa está vazia} acompanhada da explicação que a casa pode estar vazia de pessoas ou de móveis. O desempenho nesta tarefa reforça o resultado de que ambos os modelos ainda enfrentam desafios na geração de frases em que o elemento gerador de ambiguidade é um item lexical.

\textbf{Ambiguidade Sintática.}
Na análise da geração de frases com ambiguidade sintática, o ChatGPT apresentou um desempenho relativamente superior, com uma taxa de sucesso de 65\%. Ou seja, das 20 frases geradas, 13 efetivamente incorporaram ambiguidade sintática por meio de complementos sintáticos, adjuntos adverbiais ou adnominais ambíguos, demonstrando que ela conseguiu aprender tais padrões de forma mais satisfatória. Por exemplo, na frase \enquote{Ela viu o homem com o telescópio,} o ChatGPT explorou a ambiguidade gerada pelo complemento sintático \enquote{com o telescópio,} permitindo interpretações tanto de ela ter utilizado um telescópio para ver o homem quanto de o homem estar com um telescópio quando foi visto. Ela também utilizou ambiguidade oriunda de adjuntos adverbiais e adnominais, como em \enquote{A mãe deu um presente para a filha com uma fita bonita,} na qual o adjunto \enquote{com uma fita bonita} possibilita interpretações sobre o presente ter uma fita bonita ou a filha estar com uma fita bonita. Estes são os casos clássicos de ambiguidade sintática, amplamente descritos na literatura e explicados a partir dos princípios da aposição mínima e da aposição local \cite{maiadimensoes}, com descrições no português brasileiro \cite{maia2003processamento, maia2004compreensao, brito2013processamento}.

Apesar dos resultados satisfatórios obtidos na geração das frases, foram encontradas inconsistências na interpretação do elemento gerador da ambiguidade em 7 das 20 frases, mesmo quando a ambiguidade sintática estava de fato presente. Além disso, ela gerou 7 frases sem identificação de qualquer tipo de ambiguidade pelos especialistas humanos. Em 20\% das frases ela personificou elementos inanimados, prejudicando a interpretação correta. Por exemplo, na frase \enquote{Ele viu o pássaro do vizinho com binóculos.}, o ChatGPT colocou o adjunto corretamente, gerando uma sentença com potencial de ser sintaticamente ambígua, no entanto, o conhecimento de mundo de que pássaros não utilizam binóculo barra a possibilidade de haver ambiguidade para humanos.

Embora a nossa comunicação seja potencialmente ambígua, a ambiguidade não parece ser um problema, pois quando há interferência, a ambiguidade é resolvida com o esclarecimento, reparo ou correção. A IA atua como o analista, como explicam \cite{freitag2021gramatica}, que dispõem de um grande conjunto de dados desprovidos de contexto, e que portanto são propensos a gerar ambiguidade na compreensão, mas que só existem do ponto de vista da IA.

O Gemini demonstrou um conhecimento ainda mais limitado dos padrões geradores de ambiguidade sintática, gerando apenas 5 das 20 frases com adjuntos adnominais e adverbiais como fontes de ambiguidade, resultando em uma taxa de acertos de apenas 25\%. Das 15 frases erradas, 11 não apresentam nenhum tipo de ambiguidade, como em \enquote{O homem comprou o livro que estava na prateleira,} em que erroneamente ela atribuiu ambiguidade à expressão \enquote{na prateleira.}, algo que um ser humano não faria. Embora seja possível por conta dos princípios da aposição mínima e da aposição local \cite{brito2013processamento, maiadimensoes}. Além disso, em três sentenças, o Gemini confundiu o uso de palavras polissêmicas e homônimas, classificando erroneamente a ambiguidade lexical como sintática. Também é notável que, ao contrário do ChatGPT, o Gemini não incorporou a personificação de elementos inanimados em suas gerações.

Em 80\% das frases geradas, as explicações fornecidas pelo Gemini foram incoerentes demonstrando que o modelo ainda não consegue explicar corretamente a causa da ambiguidade, como na frase \enquote{O homem viu a mulher na janela,} onde atribuiu a ambiguidade ao verbo \enquote{ver,} ignorando que a verdadeira fonte era o adjunto \enquote{na janela,} permitindo que o homem ou a mulher estivessem na janela.

\textbf{Ambiguidade Semântica.}
É importante ressaltar que a ambiguidade semântica, em determinados referenciais teóricos, pode se assemelhar à ambiguidade lexical ou, em alguns casos, não é reconhecida como uma categoria distinta, resultando em uma linha tênue de separação entre ela e outros tipos de ambiguidade \cite{zavaglia2003ambiguidade}. Contudo, os resultados gerados pelo ChatGPT e pelo Gemini não se manifestam apenas na mistura de diferentes padrões de ambiguidade. Em vez disso, destaca-se na geração de frases que carecem de qualquer ambiguidade, as quais, foram consideradas como possuidoras de ambiguidade semântica. O ChatGPT, por exemplo, produziu 11 frases que não apresentaram ambiguidade para os juízes humanos, enquanto o Gemini gerou 16 frases nessas condições. As demais sentenças, em sua maioria, foram geradas com base em padrões de adjunto ambíguo ou, residindo apenas em elementos lexicais.

Ambos os modelos geraram frases semelhantes ou idênticas às produzidas para ambiguidade sintática. O ChatGPT gerou a sentença \enquote{Ela viu o homem com o telescópio.}, e o Gemini, \enquote{A menina viu o homem com o binóculo}. Ambas continham adjunto ambíguo, e padrões similares foram reproduzidos quando solicitadas frases com ambiguidade sintática, evidenciando a falta de distinção clara entre os dois tipos de ambiguidade por parte dessas versões dos modelos.

\section{Conclusão}\label{sec-conclusao}

Nossos resultados indicaram melhorias significativas entre os diferentes modelos, assim como diversas vantagens e limitações. Nesse sentido, nosso trabalho apresenta as seguintes contribuições: 1) um conjunto de dados formatado para testar a ambiguidade linguística em modelos de linguagem natural no português brasileiro, que até o nosso conhecimento é o primeiro proposto na literatura; 2) é o primeiro trabalho a informar à comunidade científica sobre as limitações do ChatGPT e do Gemini em compreender fenômenos linguísticos complexos, como a ambiguidade na língua portuguesa; 3) apresentar uma metodologia para avaliar esses modelos quanto ao fenômeno da ambiguidade; e 4) demonstrar, por meio de resultados qualitativos e quantitativos, qual dos dois modelos lida melhor com esses fenômenos linguísticos.

A análise do fenômeno de ambiguidade linguística nos modelos instrucionais ChatGPT e Gemini, cujas versões 3.5 e Bard, respectivamente, foram submetidos a quatro tarefas referentes à detecção, tipificação, desambiguação e geração de frases ambíguas. Os resultados obtidos mostraram que ambiguidade linguística ainda é um grande desafio para estas versões de modelos de processamento de linguagem natural, demandando ainda estudos e implementações para o aprimoramento.

Os modelos apresentaram baixa acurácia e baixo desempenho em praticamente todas as tarefas testadas. Na detecção de ambiguidade, o ChatGPT conseguiu uma acurácia de 28,75\% e o Gemini 49,58\%. Os modelos também apresentaram uma superinterpretação de sentenças não ambíguas, detectando e desambiguando frases que não tinham qualquer tipo de ambiguidade e que seres humanos facilmente conseguem interpretar apenas um sentido nas sentenças. Os melhores resultados obtidos ocorreram na tarefa de desambiguação e classificação onde a ambiguidade residia apenas no item lexical, demonstrando que é o tipo de ambiguidade em que os modelos tem mais facilidade para lidar. 

Algo que chamou a atenção foi que apesar da maior facilidade em lidar com ambiguidade lexical, os modelos tiveram o pior desempenho na geração de frases com ambiguidade desse tipo. Algo similar ocorreu com as frases de categoria semântica, em que os modelos confundiram com os padrões de ambiguidade estrutural ou geraram a maioria das frases sem ambiguidade. Por outro lado, ocorreu um melhor desempenho na geração de frases com ambiguidade sintática, porém ainda com uma interpretação errada sobre a origem da ambiguidade em várias frases, demonstrando que os modelos conseguiram gerar algumas frases corretamente mas ainda não conseguem explicar com clareza as causas da ambiguidade sintática. Outro ponto que chamou a atenção foi a tendência da ChatGPT em personificar alguns elementos inanimados para atribuir ambiguidade as frases, algo que um ser humano jamais faria. 

De modo geral, os resultados mostram que estas versões dos modelos instrucionais ainda estão distantes de emular plenamente a capacidade cognitiva dos seres humanos, não só envolvendo a relação entre linguagem e identidade social \cite{freitag2021preconceito}, mas também o uso na interação social, o que requer a compreensão de ambiguidades. No entanto, os resultados também sinalizam um progresso inicial na compreensão e aquisição do senso comum a respeito de como a linguagem humana funciona e reiteram a importância dos estudos descritivos em línguas ainda com poucos recursos, como é o caso do português \cite{finger2021inteligencia}, para aprimoramento.

%Bibliography

%\printbibliography\label{sec-bib}
%\bibliography{sections/article}

\bibliographystyle{unsrtnat}
\bibliography{sections/article}  %%% Uncomment this line and comment out the ``thebibliography'' section below to use the external .bib file (using bibtex) .





% \end{thebibliography}

\appendix 
\section{Apêndice}
\label{sec-apendice}

\begin{table}[!htb]
\centering
\begin{small}
\renewcommand{\arraystretch}{1.5}
\begin{tabular}{p{0.4\textwidth} p{0.4\textwidth} p{0.1\textwidth}}
\hline
\textbf{Sentenças} & \textbf{} & \textbf{Categoria} \\
\hline
João foi à mangueira, assim como Maria. & João quer mangas, assim como Maria. & L \\
João fez uma rezinha. & Isso não é legal! & L \\
Essa dama é linda! & Que gato! & L \\
Ana vai ao banco. & Acende logo! & L \\
O papel foi bem feito. & Ele está com aquela matraca. & L \\
A carteira foi danificada. & Ela estava perto da mangueira. & L \\
O homem esperava no banco. & Gostamos de rosa. & L \\
Eu gosto de damas. & Era o ponto certo. & L \\
Pedi um prato ao garçom. & Ficamos sem rede. & L \\
Pegue a pilha, por favor. & A rede caiu. & L \\
Minha mãe e minha irmã ficaram chateadas depois que ela gritou com ela. & Ela não gosta da amiga dela. & SE \\
João falou comigo e com sua mãe. & Maria e sua mãe falaram comigo. & SE \\
Ele perguntou para ela se estava bem. & Ana me contou um segredo sobre ela. & SE \\
Ele não sabia que era o dia do seu aniversário. & O professor dele escreveu várias coisas em seu caderno. & SE \\ 
O Estado deve ajudar o povo para que ele prospere. & O padre bateu o carro dele. & SE \\
Paulo não entrou na sua casa. & Ela viu o ônibus passar na rua dela. & SE \\ 
O policial prendeu o bandido em sua casa. & Para ele ficar satisfeito, o chefe preparou um prato para o convidado. & SE \\
O homem matou seu tigre. & A professora proibiu que o aluno utilizasse seu dicionário. & SE \\
Ele a viu com a sua amiga. & A moça colocou sua mão na tinta. & SE \\
José comprou pão para Maria perto de sua casa. & Os empregados se revoltaram contra os superiores por causa dos seus salários. & SE \\
Ele saiu da loja de carro. &
Joana falou com a Maria brava. & SI \\
Carlos observou Tiago malhando. & Ele viu a moça com um binóculo. & SI \\
Abandonei meu irmão contrariado. & Soube do emprego novo de Janaína no restaurante. & SI \\
O menino viu o incêndio do prédio. & Pedro me mandou um cartão postal de Veneza. & SI \\
Idosa é presa por matar homem em crise na avenida. & A filha ligou para a mãe que tinha batido o carro. & SI \\
Policial prende criminoso com arma de brincadeira. & Eu li sobre a greve dos estudantes na universidade. & SI \\
Eu avisei à Júllia que estava atrasada. & Ricardo alimentou o pássaro cansado. & SI \\
O jogador o viu de tênis. & A mãe pegou o bebê chorando. & SI \\
A moça viu o rapaz andando com a amiga na rua. & Homens e mulheres inteligentes alcançam sucesso. & SI \\
Vendo carros e caminhões usados. & Ele foi atrás do táxi apressado. & SI \\

\hline
\end{tabular}
\end{small}
\caption{Conjunto de sentenças com ambiguidade por nós desenvolvido. As categorias foram atribuídas de acordo com o referencial teórico da seção \ref{sec-referencial-teorico}.}
\label{dataset_frases_ambiguas}
\end{table}

\begin{table}[!htb]
\centering
\begin{small}
\renewcommand{\arraystretch}{1.5}
\begin{tabular}{p{0.4\textwidth} p{0.4\textwidth}}
\hline
\textbf{Sentenças} \textbf{} \\
\hline
Michel Teló tem uma sorveteria. & Eu cavo com a pá. \\
O lápis ficou desesperado quando viu que o papel acabou. & O padre bateu o carro dando ré. \\
João foi à árvore da mangueira, assim como Maria. & João e Maria gostam da fruta que se chama manga. \\
João engatou a ré. & Viajem à praia vocês! \\
Viajem vocês! & Números não é uma boa temática para filmes. \\
O homem, que ama sertanejo, estava ouvindo arrocha. & A conduta dele é inadequada, porque ele vende drogas. \\
João ganhou a partida de Canastra com uma dama na mão. & O gato fez cocô fora da caixinha de areia. \\
Eu gosto de damas, são tão educadas. & Ana vai ao banco sacar dinheiro. \\
Acende logo! Eu quero cozinhar! & Ascenda e tenha prestígio. \\
Pedi o prato principal ao garçom, era filé! & Ontem ficamos sem rede, não dava mais pra navegar na internet. \\
A rede caiu, ninguém mais balançou! & A minha mãe, que estava brava, perguntou que horas eu cheguei. \\
Carlos, que estava malhando, observou Thiago. & Ele, por meio de um binóculo, viu a moça. \\
Proibido ultrapassar. & Janaína começará a trabalhar num restaurante. \\
O bandido foi preso em flagrante pelo policial. & Pegaram emprestado o livro muito culto dele. \\
Enquanto o menino estava no prédio, viu um incêndio. & Em Veneza, Pedro comprou um cartão postal pra mim. \\
A idosa estava em uma crise e matou, na avenida, o homem. & A filha, após sofrer um acidente de carro, ligou para a mãe. \\
Um policial, portando uma arma de brinquedo, prende criminoso. & Na universidade, os estudantes faziam uma greve, que eu vi. \\
Eu avisei atrasadamente à Júlia. & Como o pássaro estava cansado, Ricardo o alimentou. \\
O Estado deve ajudar o povo, que prosperará. & O tigre matou o homem. \\
Falei que estava passando mal à chefe. & O chefe preparou um prato para que o convidado ficasse satisfeito. \\
A filha gritou com a mãe e ambas ficaram chateadas. & Ela não gosta daquela mulher. \\
João falou comigo depois de falar com a mãe pelo telefone. & Conversei com José e a mãe dele, a qual o acompanhava. \\
Ele foi embora da loja dirigindo. & Ela estava doente e avisou o chefe. \\
Ele perguntou se ela estava bem. & Ana me contou um segredo. \\
Ele não sabia que já era o dia do próprio aniversário. & O professor fez várias correções no caderno do aluno. \\
Ele está conversando com aquela mulher. & Paulo não foi até a universidade. \\
Ela não viu o ônibus que passou. & A porta foi danificada. \\
Ela estava perto da árvore. & Pegue as pilhas de roupas, por favor. \\
Gostamos da cor rosa. & O homem esperava sentado. \\
Ela é profissional em tênis de mesa. & Era o ponto de ônibus certo. \\

\hline
\end{tabular}
\end{small}
\caption{Conjunto de sentenças sem ambiguidade por nós desenvolvido.}
\label{dataset_frases_nao_ambiguas}
\end{table}

\begin{table}[!htb]
\centering
\begin{small}
\renewcommand{\arraystretch}{1.8}
\begin{tabular}{p{0.1\textwidth}
    p{0.1\textwidth}
    p{0.2\textwidth}
    p{0.1\textwidth}
    p{0.2\textwidth}
    p{0.1\textwidth}}
\hline
\textbf{modelo} & \textbf{categoria solicitada} & \textbf{sentença} & \textbf{categoria real} & \textbf{sentença} & \textbf{categoria real}\\
\hline

ChatGPT & lexical & O pássaro voou em direção à árvore mais alta. & sem ambiguidade & Ele viu o filme no cinema com os óculos escuros. & sintática\\

ChatGPT & lexical & A fuga do preso foi rápida e espetacular. & sintática & Ela encontrou uma bela maçã na floresta. & sem ambiguidade\\

ChatGPT & lexical & O livro estava na mesa do professor. & sem ambiguidade & O jogador ferido foi substituído por um novato. & sintática\\

ChatGPT & lexical & Ele comprou uma camisa azul e uma vermelha. & sem ambiguidade & O médico recomendou repouso e exercício. & sintática\\

ChatGPT & lexical & A aula de biologia foi sobre células e tecidos. & lexical & O rapaz correu para a casa da avó. & semântica\\

ChatGPT & lexical & Ela pintou a sala com as mãos. & sem ambiguidade & O barco ancorou na ilha deserta. & sem ambiguidade\\

ChatGPT & lexical & O cachorro late quando o carteiro chega. & sem ambiguidade & O guarda fechou a porta do parque. & sintática\\

ChatGPT & lexical & Ele comeu o bolo com gosto. & sintática & O homem viu o acidente com os próprios olhos. & sem ambiguidade\\

ChatGPT & lexical & O carro saiu da garagem com dificuldade. & sem ambiguidade & O escritor escreveu um romance emocionante. & sem ambiguidade\\

ChatGPT & lexical & O cantor entrou no palco com uma guitarra. & sintática & A mãe preparou a comida com amor. & lexical\\

Bard & lexical & O cachorro do vizinho mordeu a criança. & lexical & O homem comprou uma casa no campo. & sem ambiguidade\\

Bard & lexical & O advogado defendeu o criminoso. & sem ambiguidade & O soldado matou o inimigo. & sem ambiguidade\\

Bard & lexical & O médico operou o paciente. & sem ambiguidade & O professor ensinou a classe. & semântica\\

Bard & lexical & O juiz sentenciou o réu. & sem ambiguidade & O político falou para a multidão. & sem ambiguidade\\

Bard & lexical & O artista pintou um quadro. & sem ambiguidade & O escritor escreveu um livro. & sem ambiguidade\\

Bard & lexical & O carro preto passou por nós. & sem ambiguidade & A menina gorda está sentada ali. & sem ambiguidade\\

Bard & lexical & A casa está vazia. & pragmática & A rua está molhada. & sem ambiguidade\\

Bard & lexical & O livro está fechado. & sem ambiguidade & A porta está aberta. & pragmática\\

Bard & lexical & O copo está cheio. & sem ambiguidade & A mesa está limpa. & pragmática\\

Bard & lexical & O homem velho está dormindo na rua. & sem ambiguidade & A cama está arrumada. & sem ambiguidade\\

\hline
\end{tabular}
\end{small}
\caption{Conjunto de sentenças com ambiguidade lexical geradas pelos modelos de linguagem no experimento 4. A coluna de categoria real indica a classificação da sentença conforme a percepção humana da ambiguidade. Sentenças categorizadas como sem ambiguidade são aquelas em que seis seres humanos não detectaram qualquer ambiguidade. As demais categorias foram atribuídas com base no referencial teórico adotado neste trabalho.}
\end{table}

\begin{table}[!htb]
\centering
\begin{small}
\renewcommand{\arraystretch}{1.5}
\begin{tabular}{p{0.1\textwidth}
    p{0.1\textwidth}
    p{0.2\textwidth}
    p{0.1\textwidth}
    p{0.2\textwidth}
    p{0.1\textwidth}}
\hline
\textbf{modelo} & \textbf{categoria solicitada} & \textbf{sentença} & \textbf{categoria real} & \textbf{sentença} & \textbf{categoria real}\\
\hline

ChatGPT & sintática & Ela viu o homem com o telescópio. & sintática & Ele trouxe o livro para o amigo do professor. & sem ambiguidade\\

ChatGPT & sintática & O rapaz encontrou o carro dirigindo na estrada. & sintática & A professora elogiou o aluno com boas notas. & sintática\\

ChatGPT & sintática & A mãe deu um presente para a filha com uma fita bonita. & sintática & O cachorro mordeu o homem com a camisa rasgada. & sintática\\

ChatGPT & sintática & O pintor viu o quadro com os olhos cansados. & sintática & O jogador marcou um gol com a camisa amarela. & sintática\\

ChatGPT & sintática & O médico disse ao paciente que o tratamento seria difícil devido à sua saúde frágil. & sintática & O rapaz encontrou a carteira perdida no chão do parque. & sem ambiguidade\\

ChatGPT & sintática & Vi o homem com o telescópio. & sintática & Ele viu o pássaro do vizinho com binóculos. & sintática\\

ChatGPT & sintática & Eu gosto de estudar mais do que meu irmão. & sintática & Ela viu o homem com a luneta sentado na varanda. & sintática\\

ChatGPT & sintática & O cachorro do meu amigo late durante a noite. & sintática & Maria ama sua irmã mais do que eu. & sintática\\

ChatGPT & sintática & Comi a pizza com queijo. & sintática & Vi a fotografia da Maria com o cachorro. & sintática\\

ChatGPT & sintática & Ele quebrou a janela com a bola de beisebol jogando no quintal. & sintática & O professor disse ao aluno que a matéria era fácil. & sem ambiguidade\\

Bard & sintática & A menina viu o cachorro do vizinho correndo pela rua. & sintática & O homem levou a esposa ao médico. & semântica\\

Bard & sintática & O menino caiu da árvore e quebrou o braço. & semântica & A mulher comprou o vestido verde na loja. & sem ambiguidade\\

Bard & sintática & O homem viu a mulher na janela. & semântica & O gato comeu o rato que estava no armário. & sem ambiguidade\\

Bard & sintática & O menino jogou a bola para o cachorro. & semântica & A menina estava brincando com a boneca quando a mãe chegou. & semântica\\

Bard & sintática & O homem comprou o livro que estava na prateleira. & sem ambiguidade & O menino tirou a roupa e foi tomar banho. & semântica\\

Bard & sintática & O cachorro do meu tio só dá trabalho. & sintática & Carla falou pro marido que tinha chegado. & sintática\\

Bard & sintática & Eu fiquei te esperando no banco. & lexical & Ela puxou a cadeira e acabou quebrando o braço. & semântica\\

Bard & sintática & A galinha brincava com uma xícara até que quebrou a asa. & semântica & O guarda deteve o suspeito em sua casa. & sintática\\

Bard & sintática & O cachorro comeu o osso e o menino. & sem ambiguidade & O homem comprou um carro novo e uma casa. & sem ambiguidade\\

Bard & sintática & A mulher estava com medo do cachorro. & sem ambiguidade & O menino estava andando de bicicleta na rua. & sem ambiguidade\\

\hline
\end{tabular}
\end{small}
\caption{Conjunto de sentenças com ambiguidade sintática que foram geradas pelos modelos de linguagem no experimento 4. A coluna de categoria real indica a classificação da sentença conforme a percepção humana da ambiguidade. Sentenças categorizadas como sem ambiguidade são aquelas em que seis seres humanos não detectaram qualquer ambiguidade. As demais categorias foram atribuídas com base no referencial teórico adotado neste trabalho.}
\end{table}

\begin{table}[!htb]
\centering
\begin{small}
\renewcommand{\arraystretch}{1.8}
\begin{tabular}{p{0.1\textwidth}
    p{0.1\textwidth}
    p{0.2\textwidth}
    p{0.1\textwidth}
    p{0.2\textwidth}
    p{0.1\textwidth}}
\hline
\textbf{modelo} & \textbf{categoria solicitada} & \textbf{sentença} & \textbf{categoria real} & \textbf{sentença} & \textbf{categoria real}\\
\hline

ChatGPT & semântica & Ela viu o homem com o telescópio. & sintática & O banco estava vazio, exceto por um único ocupante. & sem ambiguidade\\

ChatGPT & semântica & Ele pegou a maçã verde do chão. & sem ambiguidade & Vi o pássaro da janela enquanto dirigia. & sem ambiguidade\\

ChatGPT & semântica & O jogador atirou a bola para o goleiro. & sem ambiguidade & Ele encontrou a carta de Maria na gaveta. & sem ambiguidade\\

ChatGPT & semântica & A aula foi interessante para os alunos cansados. & sem ambiguidade & A luz do sol iluminava a sala escura. & sem ambiguidade\\

ChatGPT & semântica & A pele do caçador era macia. & lexical & O cachorro viu o gato no telhado com binóculos. & sintática\\

ChatGPT & semântica & Ela viu o homem com o telescópio. & sintática & Ele encontrou a mulher de chapéu na festa. & sintática\\

ChatGPT & semântica & O pato é bom para comer. & sintática & As crianças viram o filme no parque. & sem ambiguidade\\

ChatGPT & semântica & O ladrão roubou o banco com uma máscara. & sintática & Ele leu o livro com os óculos escuros. & sem ambiguidade\\

ChatGPT & semântica & O cachorro viu o gato na varanda. & sintática & Ela pintou o quadro com a mão. & sintática\\

ChatGPT & semântica & O jogador ferido saiu do campo mancando. & sem ambiguidade & A situação melhorou após a conversa com o terapeuta. & sem ambiguidade\\

Bard & semântica & A menina viu o homem com o binóculo & sintática & O homem viu a mulher com o cachorro & sintática\\

Bard & semântica & O menino quebrou o braço da cadeira. & sem ambiguidade & A casa do João é azul & sem ambiguidade\\

Bard & semântica & O livro do menino está na mesa & sem ambiguidade & O homem está morto & sem ambiguidade\\

Bard & semântica & O menino está com fome. & sem ambiguidade & A mulher está cansada & sem ambiguidade\\

Bard & semântica & O cachorro está feliz & sem ambiguidade & O gato está dormindo & sem ambiguidade\\

Bard & semântica & O homem está procurando um cachorro. & sem ambiguidade & O menino foi ao banco. & lexical\\

Bard & semântica & O guarda prendeu o suspeito em sua casa. & semântica & O professor deu uma bronca na aluna. & sem ambiguidade\\

Bard & semântica & O carro está parado no sinal vermelho. & sem ambiguidade & O homem está com fome. & sem ambiguidade\\

Bard & semântica & A mulher está com frio. & sem ambiguidade & O menino está com medo. & sem ambiguidade\\

Bard & semântica & A menina está com alegria. & sem ambiguidade & O homem está com raiva. & sem ambiguidade\\

\hline
\end{tabular}
\end{small}
\caption{Conjunto de sentenças com ambiguidade semântica que foram geradas pelos modelos de linguagem no experimento 4. A coluna de categoria real indica a classificação da sentença conforme a percepção humana da ambiguidade. Sentenças categorizadas como sem ambiguidade são aquelas em que seis seres humanos não detectaram qualquer ambiguidade. As demais categorias foram atribuídas com base no referencial teórico adotado neste trabalho.}
\end{table}

\end{document}